\documentclass[draft]{agujournal2019}
\usepackage{url} 
\usepackage[inline]{trackchanges} 
\usepackage{soul}
\usepackage{amsmath}   
\usepackage{amssymb}   
\draftfalse
\begin{document}

%
%


\title{Skillful High-Resolution Ensemble Precipitation Forecasting with an Integrated Deep Learning Framework}

%
%




\authors{Shuangshuang He\affil{1}, Hongli Liang\affil{1}, Yuanting Zhang\affil{1, *}, Xingyuan Yuan\affil{1}}


\affiliation{1}{ColorfulClouds Technology Co.,Ltd., Beijing, China.}




\correspondingauthor{Yuanting Zhang}{zhangyuanting@caiyunapp.com}



\begin{keypoints}
\item We proposed a physics-inspired hybrid deep learning framework for high-resolution ensemble precipitation forecasting.
\item The framework combines a deterministic model to represent mesoscale evolution and a probabilistic model to capture uncertainties at a finer scale. 
\item The model demonstrates improved representation of extreme precipitation and accurate real-time forecasts.
\end{keypoints}

%
%

%
%


\begin{abstract}
High-resolution precipitation forecasts are crucial for providing accurate weather prediction and supporting effective responses to extreme weather events. Traditional numerical models struggle with stochastic subgrid-scale processes, while recent deep learning models often produce blurry results. To address these challenges, we propose a physics-inspired deep learning framework for high-resolution (0.05\textdegree{} $\times$ 0.05\textdegree{}) ensemble precipitation forecasting. Trained on ERA5 and CMPA high-resolution precipitation datasets, the framework integrates deterministic and probabilistic components. The deterministic model, based on a 3D SwinTransformer, captures average precipitation at mesoscale resolution and incorporates strategies to enhance performance, particularly for moderate to heavy rainfall. The probabilistic model employs conditional diffusion in latent space to account for uncertainties in residual precipitation at convective scales. During inference, ensemble members are generated by repeatedly sampling latent variables, enabling the model to represent precipitation uncertainty. Our model significantly enhances spatial resolution and forecast accuracy. Rank histogram shows that the ensemble system is reliable and unbiased. In a case study of heavy precipitation in southern China, the model outputs align more closely with observed precipitation distributions than ERA5, demonstrating superior capability in capturing extreme precipitation events. Additionally, 5-day real-time forecasts show good performance in terms of CSI scores.
\end{abstract}

\section*{Plain Language Summary}
Deep learning models often produce blurring results and reduced intensity in precipitation predictions. We developed a deep learning framework that achieves skillful high-resolution precipitation forecasting in this study. Our model combines deterministic and probabilistic components to capture large-scale precipitation patterns and the uncertainties associated with smaller, intense convective activity. Using coarse-resolution (0.25\textdegree{}) atmospheric variables (temperature, pressure, humidity, geopotential and wind) from NWP or data-driven models as inputs, the model is able to generate high-resolution (0.05\textdegree{}) precipitation predictions. The evaluation results show that our ensemble forecasting system exhibits unbiased results and demonstrates superior performance in capturing extreme precipitation events. Additionally, we implemented a real-time forecasting system using ECMWF open data, which further demonstrates the model’s robustness and reliability.

%
%

%


%
%
%
%

\section{Introduction}
Accurate prediction of precipitation, particularly at high spatial and temporal resolutions, is critical for numerous applications, including flood forecasting, water resource management, and disaster mitigation \cite{trenberth2011}. However, precipitation forecasting remains one of the most challenging tasks in atmospheric sciences due to the inherent complexity and stochastic nature of small-scale atmospheric processes \cite{bauerQuietRevolutionNumerical2015}.

In traditional numerical weather prediction (NWP) models, precipitation is often parameterized due to their inability to explicitly resolve small-scale convective processes and microphysical processes \cite{Molinari1992Review, iorio2004effects, skamarock2005description}. These parameterization schemes attempt to approximate the effects of unresolved processes on grid-scale fields, but they introduce significant biases and errors \cite{tiedtke1989comprehensive, kain1993convective, molteni1996ecmwf, palmer2005representing}. As a result, the accuracy of traditional models in predicting small-scale precipitation events, such as convective storms or heavy rainfall, is often limited \cite{zhao2021which, allen2022model}. Moreover, running sophisticated NWP models at high spatial resolutions to better resolve precipitation processes comes with immense computational costs, making real-time operational forecasting a challenge \cite{tang2013benefits}. This trade-off between model resolution and computational cost remains a significant limitation of traditional weather prediction approaches.

Recent advancements in deep learning offer promising alternatives to improve both the accuracy and efficiency of weather forecasting. Models such as Pangu-Weather, GraphCast, and Fuxi have demonstrated the potential of data-driven approaches to produce accurate weather forecasts at a fraction of the computational cost of traditional NWP models \cite{biAccurateMediumrangeGlobal2023, lam2023learning, chenFuXiCascadeMachine2023}. These models utilize large datasets and leverage architectures like transformers or graph neural networks (GNN) to capture complex atmospheric patterns from data, making it possible to generate faster predictions without the need for explicit physical modeling. Despite these advancements, challenges persist in precipitation forecasting, particularly in maintaining spatial detail over longer forecast lead times. Many deep learning models tend to produce overly smoothed and weaker outputs, losing the necessary detail required for accurate short-term precipitation forecasting \cite{charlton-perez2024, liu2024}. Furthermore, models trained on coarser-resolution data (e.g., 0.25° spatial resolution and 6-hour temporal intervals) are typically unable to capture the small-scale convective processes required for accurate precipitation prediction. 
Unlike variables such as temperature, wind, and pressure, which are directly governed by physical laws, precipitation is influenced by highly complex and stochastic microphysical processes \cite{ritchie1995implementation}, making it more challenging for deep learning models to predict. Some models, such as FourCastNet, have attempted to address this by training a separate diagnostic model using the outputs of the backbone network to predict precipitation \cite{pathak2022}. However, most of the approaches still result in overly smoothed outputs \cite{zhou2022, liuEnhancingQuantitativePrecipitation2024}.

Recently, diffusion models have emerged as a powerful probabilistic generative approach, originally achieving remarkable success in image generation \cite{hoDenoisingDiffusionProbabilistic2020, songDenoisingDiffusionImplicit2022, rombachHighResolutionImageSynthesis2022}. Diffusion models add noise gradually, transforming complex distributions into simple Gaussian forms and then generating high-fidelity image samples through a reverse process. This framework offers unique advantages in capturing complex probability distributions, making it particularly useful for handling small-scale phenomena with high uncertainty. Some studies have already begun exploring the use of diffusion models for precipitation nowcasting \cite{gaoPreDiffPrecipitationNowcasting2023, aspertiPrecipitationNowcastingGenerative2023, gongCasCastSkillfulHighresolution2024}, showing promise in capturing uncertainties associated with convective-scale precipitation.

In this study, we propose a physics-inspired deep learning framework that combines the strengths of deterministic and probabilistic modeling approaches to address the challenges of accurate high-resolution precipitation forecasting. The deterministic model captures the meso-scale evolution of precipitation, while the probabilistic component, based on latent conditional diffusion, accounts for the uncertainties associated with small-scale convective precipitation. This approach allows the model to generate ensemble forecasts, providing a more comprehensive representation of uncertainty, particularly for extreme precipitation events. Our framework takes coarse-resolution atmospheric variables, such as temperature, pressure, geopotential, humidity, and wind at a 0.25\textdegree{} resolution (from either traditional NWP or data-driven forecasts) as input and generates high-resolution precipitation forecasts at a 0.05\textdegree{} resolution, with hourly predictions.

The key contributions of this paper are:
\begin{enumerate}
\item We propose a novel deep learning framework for high-resolution ensemble precipitation forecasting, combining a deterministic model to represent meso-scale evolution and a probabilistic model that captures uncertainties at the convective scale through diffusion model in the latent space.
\item Our results demonstrate that the ensemble forecasts are nearly unbiased and the probabilistic distributions closely match the target precipitation distributions.
\item We successfully implemented a real-time forecast system for up to five days, with good performance on Critical Success Index(CSI) over the forecast horizon.
\end{enumerate}


\section{Data}
\subsection{ERA5}
For training our model, we built our datasets from ERA5 archive, the fifth generation of ECMWF reanalysis dataset \cite{hersbach2020era5}. ERA5 is generated using ECMWF’s Integrated Forecast System (IFS) cycle 42r1, operational for most of 2016. It employs an ensemble 4D-Var data assimilation scheme that incorporates 12-hour windows of observations from 21-09 UTC and 09-21 UTC, along with previous forecasts, allowing for a dense representation of the weather's state for each historical date and time. ERA5 assimilates high-quality global observations, making it widely regarded as the most accurate and comprehensive reanalysis archive.

We use a subset of ERA5 dataset on a 0.25\textdegree{} equiangular grid spanning 0-60\textdegree{} N, 70-140\textdegree{} E. The grid overall contains 241 × 281 grid points for latitude and longitude, respectively. 5 surface variables and 5 upper-air variables at 13 pressure levels (corresponding to the levels of the WeatherBench \cite{rasp2021data} benchmark: 50, 100, 150, 200, 250, 300, 400, 500, 600, 700, 850, 925, and 1000 hPa). The 5 upper-air atmospheric variables are geopotential(Z), temperature(T), u component of wind(U), v component of wind(V), and specific humidity(SH). Additionally, 5 surface variables are 2-meter temperature(T2M), 10-meter u wind component(U10), 10-meter v wind component(V10), mean sea level pressure(MSLP), and total precipitation(TP).

\subsection{CMPA}
A high-resolution precipitation dataset, CMPA, was utilized in this study \cite{xie2011}. Developed by the China Meteorological Administration (CMA), CMPA combines multiple data sources, including observations from 30,000 to 40,000 ground-based stations, radar-derived quantitative precipitation estimates, and satellite precipitation retrievals from FY-2E and CMORPH (CPC MORPHing technique). By merging these sources, CMPA enhances the accuracy and resolution of precipitation measurements. It provides data at a high spatiotemporal resolution of 0.01\textdegree{} $\times$ 0.01\textdegree{} and 1-hour intervals, covering the region from 15\textdegree{}N to 60\textdegree{}N and 70\textdegree{}E to 140\textdegree{}E (4500 latitude $\times$ 7000 longitude grid points).

To reduce computational burden, we applied nn.AvgPool2d with PyTorch to transform the dataset into a lower resolution of 0.05\textdegree{} $\times$ 0.05\textdegree{}.

\subsection{ECMWF real-time forecasts}
Since February 2024, ECMWF has made open real-time medium-range forecast data \cite{ECMWF2023} from the Integrated Forecasting System (IFS) available at an improved resolution of 0.25\textdegree{} $\times$ 0.25\textdegree{}. This dataset is publicly available and is well-suited for machine learning applications, providing a valuable resource for enhancing forecast accuracy. To generate high-resolution ensemble precipitation predictions, we use the forecast data from the 00 and 12 UTC initializations, spanning the next 120 hours at 3-hour intervals.

\subsection{Preprocessing}
All other variables are normalized using the mean and standard deviation except for precipitation,. Specifically, for precipitation, we first convert the 1-hour accumulated precipitation TP (mm) to reflectivity (dBZ) using the following formula \cite{marshall1948distribution}:
\begin{equation}
dBZ = 10 \times \log_{10}(200 \times TP^{1.6})
\end{equation}
Afterward, the ERA5 and CMPA precipitation data are typically normalized to the [0, 1] range using the spatial maximum averaged over time, following the method outlined by \citeA{bodnarFoundationModelEarth2024}:
\begin{equation}
dBZ^{scale} = \frac{1}{N} \sum_{t=1}^{N} \max \{ X_{t, i, j} : i = 1, \dots, H; j = 1, \dots, W \}
\end{equation}

\section{Methodology}
\subsection{Problem formulation}
\begin{figure}[htp]
    \centering
    \noindent\includegraphics[width=8cm]{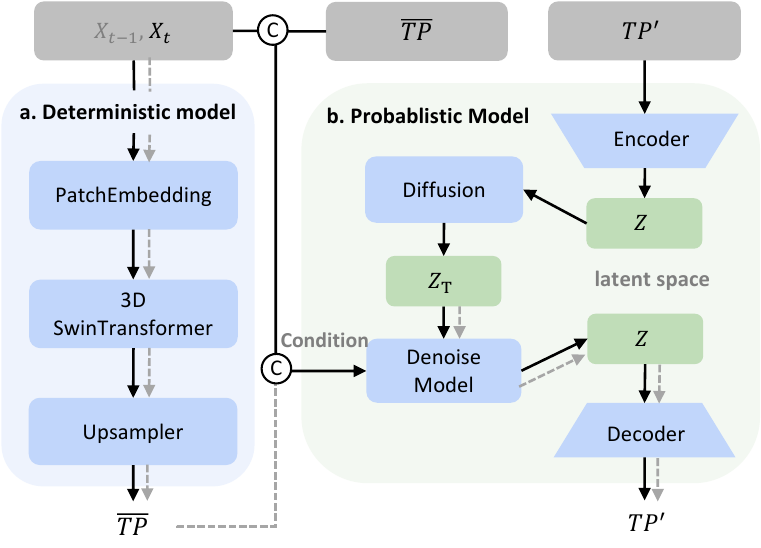}
    \caption{Overall model architeture. $X_{t-1}/X_t$ includes surface variables (T2m, U10m, V10m, MSLP) and upper-air variables at 13 pressure levels (T, U, V, SP, Z) at time $t-1/t$. For the deterministic model, the inputs are $X_{t-1}$ and $X_t$, while for the probabilistic model, the input is $X_t$. The model is trained on ERA5 data but can take forecasts from other models during inference. $\overline{TP}$ is the mean precipitation at time t, $TP'$ is the residual precipitation. $Z$ is the latent space, and $Z_T$ is the latent variable at diffusion step T. The circle with “C” represents concatenation, black arrows represent the training process, and gray dashed lines indicate the inference process.}
    \label{fig:model_arch}
\end{figure}

Inspired by the approach used in numerical weather models, where physical variables in the atmospheric governing equations can be represented as a combination of mean values and perturbations. The mean values correspond to the grid point values, and the perturbations account for sub-grid scale processes. These sub-grid processes are represented using physical parametrization schemes.
\begin{equation}
X_t = \bar{X_t} + X_t'
\end{equation}
\begin{figure}[htp]
    \centering
    \includegraphics[width=10cm]{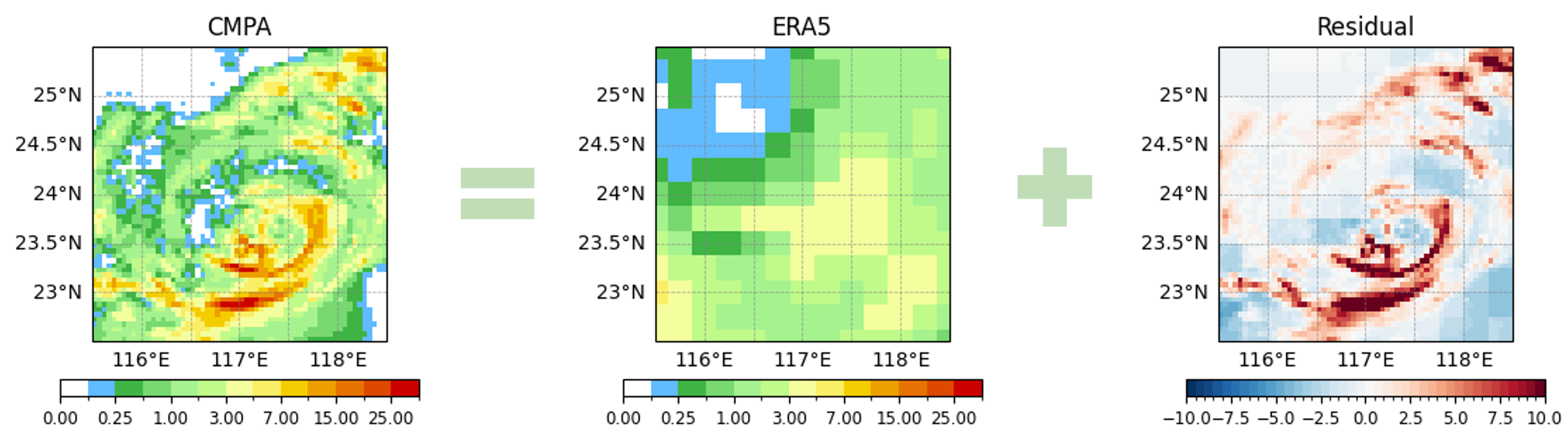}
    \caption{Decomposition of high-resolution precipitation}
    \label{fig:model_design}
\end{figure}
Similarly, we assume that high-resolution precipitation is composed of the average precipitation over a meso-scale $\overline{TP}$, plus the residual precipitation at the sub-grid scale $TP'$. We derive the residual precipitation $TP'$ from the difference between scaled CMPA and ERA5 precipitation. ERA5 precipitation was interpolated to a 0.05° grid resolution using the nearest-neighbor method to match CMPA’s resolution.
As shown in Figure~\ref{fig:model_arch}, we utilize a deterministic component for the mean precipitation and a latent diffusion component for the residual precipitation. Considering that processes on the sub-grid scale often exhibit randomness and uncertainty, we employ ensemble forecasting to capture and quantify these uncertainties. We run multiple stochastic sampling processes through the conditional diffusion model. Each sample represents a possible realization of the precipitation field, conditioned on the same input atmospheric states. 
\begin{equation}
E=\{TP_i | \overline{TP}+TP_i', i\in[1, 11]\}
\end{equation}

\subsection{Deterministic model}
\subsubsection{Model input}
The deterministic model uses four ERA5 surface variables (T2m, U10m, V10m, and MSLP, with a shape of 4 × 241 × 281) and upper air variables at 13 pressure levels (T, U, V, SP, and Z,  with a shape of 5 × 13 × 241 × 281) of two consecutive time steps ($[X_{t-1}, X_t]$) to predict 1-hour accumulated precipitation at $X_t$. In addition to these variables, we also incorporated static features such as the binary land-sea mask, geopotential at the surface, soil type, latitude and longitude in radians, as well as temporal features including the sine and cosine of the local time of day and the sine and cosine of the yearly progress. These additional features were shown to improve the diagnostic prediction of precipitation, as demonstrated in the \ref{ablationStudy}.
\subsubsection{Model architecture} \label{dtmArch}
The deterministic model architecture consists of three main components: 
\begin{enumerate}
    \item Patch embedding, which divides the input variables into small patches and embeds them into high-dimensional feature representations.
    \item 3D Swin-Transformer, which processes the embedded data by capturing both local and global features through multi-head self-attention mechanisms in spatial and temporal dimensions.
    \item Upsampler, which process the output of the 3D-Swin Transformer to produce the final precipitation prediction at the original data resolution.
\end{enumerate}

In a standard patch embedding for vision transformers, convolution is typically used to project the input data into embedded patches. However, this approach produces outputs that reflect only the linear relationships between the variables after convolution. To address this, we propose an non-linear approach: applying separate convolutions to each variable individually, followed by a GELU (Gaussian Error Linear Unit) activation. The outputs are then passed through an MLP layer and another GELU activation, ensuring that each patch integrates nonlinear information from the variables. We use a $4 \times 4 $ convolution kernel for surface variables and a $2 \times 4 \times 4$ convolution kernel for upper air variables. This results in an output with a shape of $C \times 1 \times 61 \times 71$ for the surface variables and $C \times 7 \times 61 \times 71$ for the upper-air variables. After patch embedding, these two outputs are concatenated along the channel dimension, creating a combined input with a shape of $C \times 8 \times 61 \times 71$.

The input data, with a size of $C \times 8 \times 61 \times 71$, is processed through a 3D Swin-Transformer \cite{liu2021} consisting of three layers. The first layer has 3 blocks and is followed by a patch merging operation that downsamples the data to $2C \times 8 \times 31 \times 36$. In the second layer, which contains 9 blocks, the data is further processed and then upsampled back to its original size of $C \times 8 \times 61 \times 71$ before being passed into the third layer, which also has 3 blocks. The output of the first layer is added to the output of the second layer through a skip connection. The outputs from both the first and second layers are added to the third layer’s output via skip connections. 

For generating the final output, we validated two methods. In the first method (Upsampler1), we used a 3D convolution with a kernel size of ($ 8 \times 1 \times 1 $), followed by bilinear upsampling, which doubled the height and width dimensions as a shape of C $ \times 1 \times 121 \times 141$. We then applied a 2D convolution with bilinear upsampling to restore the data to its original resolution of $ 1 \times 241 \times 281$ . In the second method (Upsampler2), we used a 3D convolution with a ($ 8 \times 1 \times 1 $) kernel and a GELU activation to reduce the height dimension to 1. Afterward, we adopted an image reconstruction technique used by SwinIR \cite{liang2021}, which includes several convolutional layers: one before upsampling, followed by others for further processing. These layers consist of LeakyReLU activations and multiple convolutions to refine the output, ultimately restoring the high-resolution data. A comparison of the results produced by these two methods can be found in the \ref{ablationStudy}.
\subsubsection{Weighted loss function}
The loss function used is the weighted L2 and SIMM (Structural Similarity Index Measure) loss, which is defined as follows:
\begin{equation}
Loss = \lambda1 * (\hat{X}_{i, j} - X_{i, j})^2 + \lambda2 * (1-SSIM(\hat{X}_{i,j}-X_{i,j}))
\end{equation}
where $\lambda1$ and $\lambda2$ are weights that balance L1 and SSIM, is set to 0.5 and 1.5 respectively. The SSIM loss \cite{ssim} is sensitive to changes in contrast, luminance, and texture, making it more effective at preserving the spatial distribution and intensity of features like heavy rainfall.

\subsection{Probabilistic model}
The probabilistic model is designed to generate residual precipitation, using the current atmospheric state plus mean precipitation as a conditional input. Two key approaches are employed in this model: Variational Autoencoder (VAE) and Conditional Diffusion.
\subsubsection{Variational autoencoder}
To improve the training efficiency of the diffusion model, we use a VAE model to encode the residual precipitation data, with dimensions $1 \times 900 \times 1400$, into a latent representation of size $16 \times 90 \times 140$. The VAE is used to encode samples from the pixel space to a continuous latent space and then decode them back to the pixel space \cite{kingma2022}. We train our autoencoder models following \citeA{rombachHighResolutionImageSynthesis2022}, which uses an adversarial manner to enhance the generative quality of the model. The encoder and decoder components of the VAE in this study are constructed as 2D convolutional networks. In the encoder, the first three layers contain two ResNet-type residual blocks and a downsampling block, each layer reduces the spatial dimensions by a factor of 2, while the final layer consists of two ResNet blocks without downsampling. Since our target size is not a power of 2, the feature maps are interpolated to the target resolution $ 90 \times 140 $. The decoder mirrors the encoder structure, using upsampling to progressively restore the spatial resolution, and interpolation to match the final output size to the input. The latent space is regularized using a Kullback–Leibler (KL) divergence loss to align the latent variables with a multivariate standard normal distribution. 

\subsubsection{Conditional latent diffusion model}
The conditional diffusion model we use is based on the Denoising Diffusion Probabilistic Model (DDPM), which progressively corrupts a data sample by adding Gaussian noise over a series of timesteps. A neural network is then trained to reverse this noising process, recovering the original data.
\begin{figure}[htp]
    \centering
    \noindent\includegraphics[width=6cm]{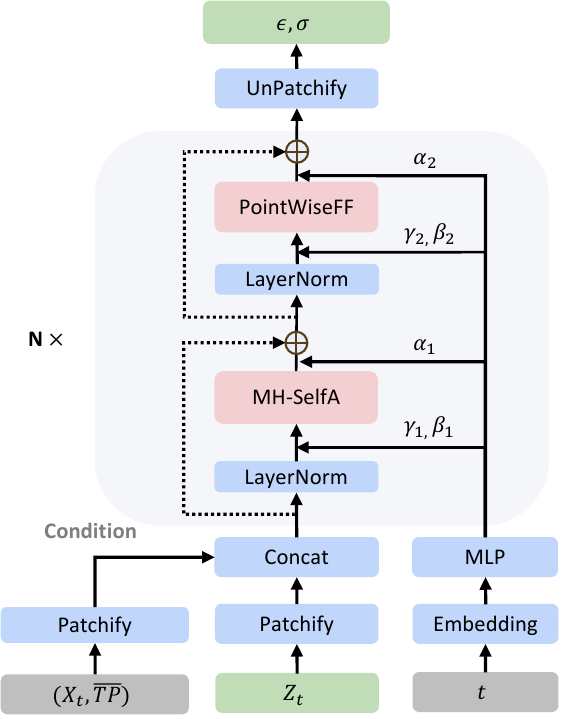}
    \caption{Conditional diffusion model architecture \cite{peeblesScalableDiffusionModels2023}. We introduce conditioning inputs by using the atmospheric state and mean precipitation.}
    \label{fig:dit}
\end{figure}
The conditional latent diffusion model training objective can be formulated as: 
\begin{equation}
\mathcal{L} = \mathbb{E}_{embedding(x), y, \epsilon \sim \mathcal{N}(0, 1), t} \left\| \epsilon - \epsilon_{\theta}(Z_t, t, cond) \right\|^2
\end{equation}

where $x$ is the condition, which is the atmospheric state, $y$ is the residual precipitation, $\epsilon$ is random noise, $t\in[1, 1000]$ is the time step of the denoising process, $z_t$ is the noisy latent-space sample at step $t$, $\epsilon_\theta$ is the denoiser and $\theta$ represents the trainable parameters of the networks. 
In our setup, we train the denoise model conditioned on ERA5 surface variables (T2m, U10m, V10m, MSLP and TP) and upper air variables at 13 pressure levels (T, U, V, SP, and Z) to guide the denoising process and improve the accuracy of the generated residual precipitation. During inference, we replace ERA5 precipitation with the predicted 1-hour precipitation from the deterministic model.

The denoising network $\epsilon_\theta$ of our model referred to the DiT (Diffusion Transformer) architecture, a transformer-based model specifically designed for diffusion processes \cite{peeblesScalableDiffusionModels2023}. DiT utilizes self-attention mechanisms to capture both global and local dependencies in the data. The conditioning variables are first processed through a 2D patch embedding as the improved approach we proposed in \ref{dtmArch}. The patch size for the conditional variables is $4 \times 4$, while the noise state has a patch size of $2 \times 2$. After patchifying, both the conditional input and the noise state have the same spatial dimensions of $45 \times 75$. The patch-embedded conditional input is then concatenated along the channel dimension with the noisy state of the diffusion model at each timestep.

Since CMPA coverage region is smaller than the area used for training the deterministic model with ERA5, the ERA5 variables are cropped to the region of 15\textdegree{}N-60\textdegree{}N and 70\textdegree{}E-140\textdegree{}E ($181 \times 241$) before being input into probabilistic model. 

\subsection{Model training and inference}
The deterministic model is trained on ERA5 data from 2001 to 2019. For the probabilistic model, the training dataset spans the years 2018-2019, with 2020 used as the validation set and 2021 reserved for testing. The deterministic model was trained for approximately 69 hours on 2 Nvidia A100 GPUs, while the probabilistic model took 38 hours to train.

During the inference phase, the probabilistic model first samples from the latent space and then uses ERA5 variables and precipitation predictions from the deterministic model as conditioning inputs. Sampling is performed using the Denoising Diffusion Implicit Models (DDIM) approach \cite{songDenoisingDiffusionImplicit2022}, with 300 sampling steps. This process is repeated for 11 iterations, generating 11 ensemble forecast members, which provides a probabilistic estimate of precipitation.
\section{Results}
\subsection{Ablation study of deterministic model} \label{ablationStudy}
Due to the tendency of deterministic models to underestimate precipitation intensity and blurriness, we adopted several methods to improve the performance of precipitation forecasting. These methods are as follows (For more details about these methods, refer to \ref{dtmArch}:

\begin{enumerate}
\item  Exp-d1: Weighted MSE and SSIM loss: MSE ensures pixel-wise accuracy, while SSIM helps the model preserve the spatial distribution and intensity of features like heavy rainfall.
\item  Exp-d2:Incorporation of static and temporal (ST) features: Enables the model better learn local and temporal variations.
\item  Exp-d3:Non-linear patch embedding: Enhances the patch embedding by applying depth-wise convolution and activation layers, allowing each patch to extract more non-linear interactions.
\item  Exp-d4:Refined upsampler: Employs additional convolutional layers in the upsampling decoder to improve the recovery of precipitation details.
\end{enumerate}

\begin{table}[htp]
\centering
\caption{Experiment Setup of Determinstic model. ST Features represents static and temporal features}
\label{tab:ablation_study}
\begin{tabular}{l l l l l}
\hline
\textbf{Experiment} & \textbf{Loss} & \textbf{ST Features} & \textbf{Embedding} & \textbf{Upsampler} \\ \hline
\textbf{baseline}    & MSE           & No                                  & Standard           & Upsampler1          \\ \hline
\textbf{exp-d1}      & MSE+SSIM      & No                                  & Standard           & Upsampler1          \\ \hline
\textbf{exp-d2}      & MSE+SSIM      & Yes                                 & Standard           & Upsampler1          \\ \hline
\textbf{exp-d3}      & MSE+SSIM      & Yes                                 & Non-Linear         & Upsampler1          \\ \hline
\textbf{exp-d4}      & MSE+SSIM      & Yes                                 & Non-Linear         & Upsampler2          \\ \hline
\end{tabular}
\end{table}

\begin{figure}[ht]
    \centering
    \includegraphics[width=14cm]{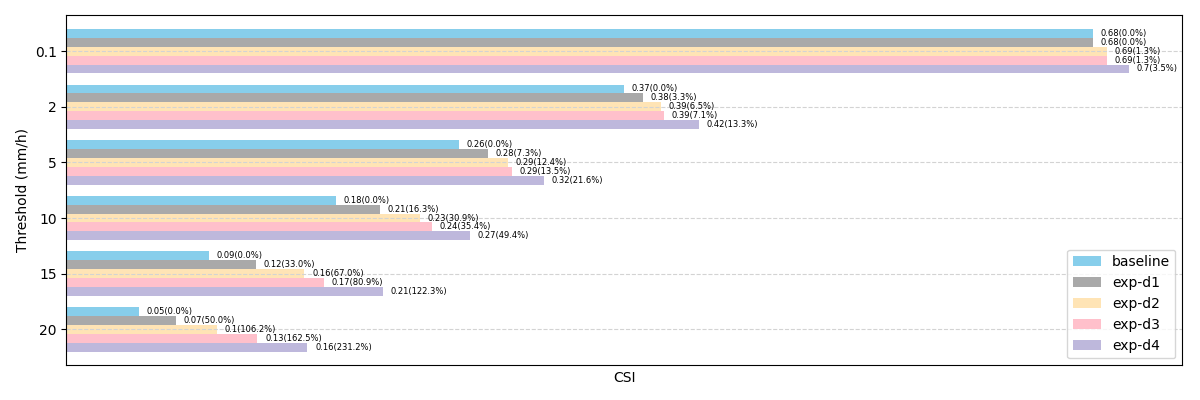}
    \caption{CSI of ablation studies against multiple precipitation thresholds, the experiment configuration referred Table~\ref{tab:ablation_study}}
    \label{fig:percent_score}
\end{figure}

The data from 2016 to 2019 were used as training dataset for the ablation study, with August, 2020 serving as the validation set.
Figure~\ref{fig:percent_score} shows the CSI scores of different experiments under various rainfall thresholds. From Figure~\ref{fig:percent_score}, it is evident that all methods improved CSI scores for precipitation, with the enhancement becoming more pronounced as the precipitation intensity increases. In exp-d1, the use of weighted SSIM and MSE loss, compared to using only MSE as the loss function, improved the CSI scores for rainfall above thresholds of 2 mm, 5 mm, 10 mm, 15 mm, and 20 mm by 3.3\%, 7.3\%, 16.3\%, 33\%, and 50\%, respectively. Building on exp-d1, we introduced static and temporal features in exp-d2. When compared to exp-d1, exp-d2 shows improvements at all thresholds, specifically by 1.3\%, 3.2\%, 5.1\%, 14.6\%, 34\%, and 56.2\%. Exp-d3 introduces nonlinear patch embedding compared to exp-d2. While the improvements are not very significant for lighter rainfall, with no noticeable gain at the 0.1 mm/h threshold and only slight improvements of 0.6\%, 1.1\%, and 4.5\% at the 2 mm, 5 mm, and 10 mm thresholds, it shows considerable gains at higher intensities, with improvements of 13.9\% and 56.3\% at the 15 mm and 20 mm thresholds, respectively. Finally, exp-d4, which employs an upsampler with more convolutional layers for feature preservation, shows significant improvements over exp-d3, with gains of 2.4\%, 6.2\%, 8.1\%, 14.0\%, 41.4\%, and 68.7\% at the respective thresholds, making it the most effective among all experiments.

\subsection{Ensemble forecast evaluation}
We conducted several evaluations using ERA5 variables as input to assess the model’s performance, focusing on the period of August 2021.
\subsubsection{Real case-August 1, 2021}
\begin{figure}[ht]
    \centering
    \noindent\includegraphics[width=14cm]{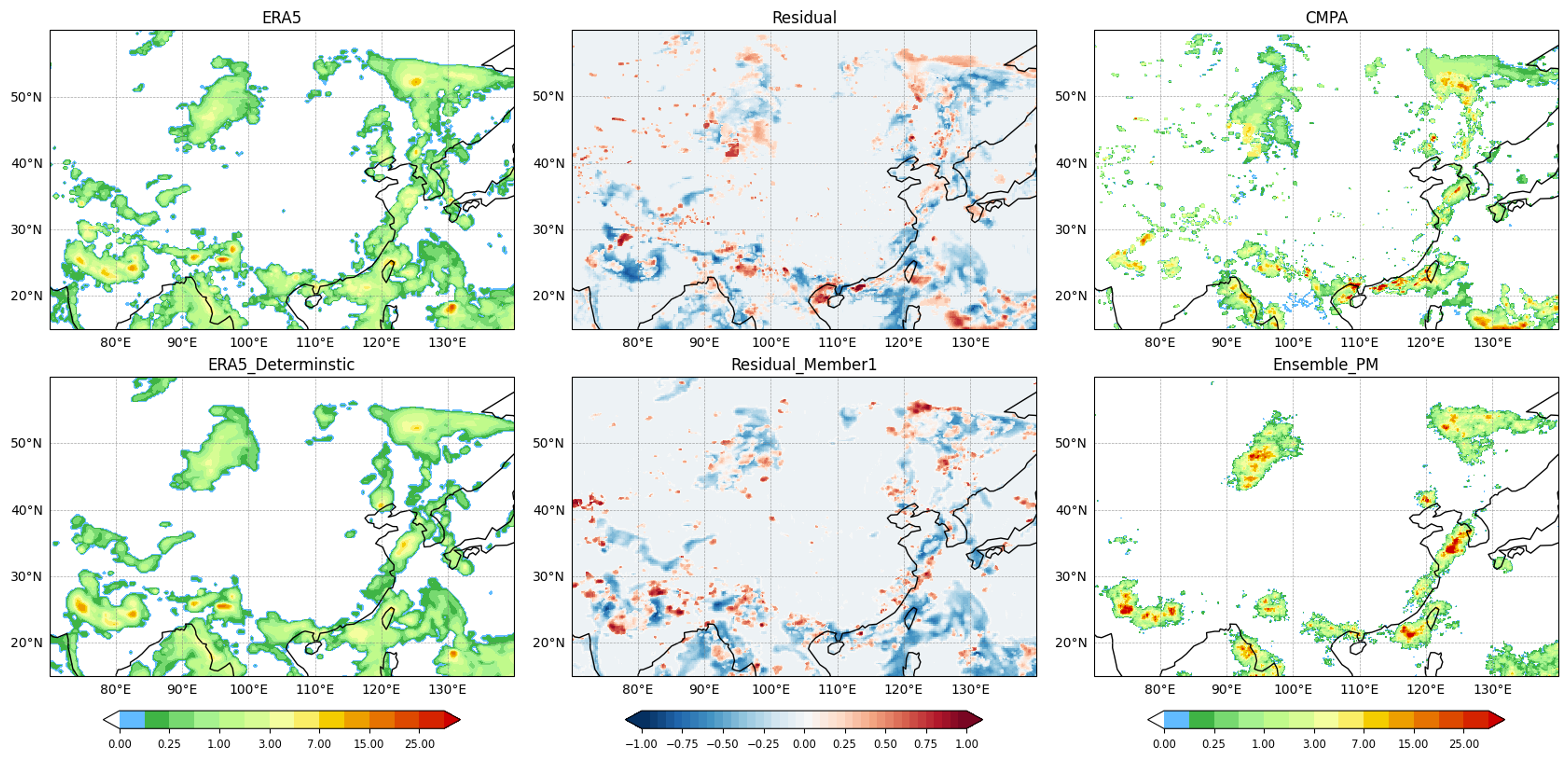}
    \caption{A case on 2021-08-01 01:00. The first row shows the ground truth, while the second row shows the predictions. The leftmost image in the second row represents the mean precipitation generated using ERA5 as input, the middle image (Residual\_Member1) represents one member of the residual precipitation, and the rightmost image (Ensemble\_PM) shows the ensemble mean precipitation obtained by probability matching \cite{ebert2001} across all ensemble members.}
    \label{fig:case1}
\end{figure}
\begin{figure}[ht]
    \centering
    \noindent\includegraphics[width=14cm]{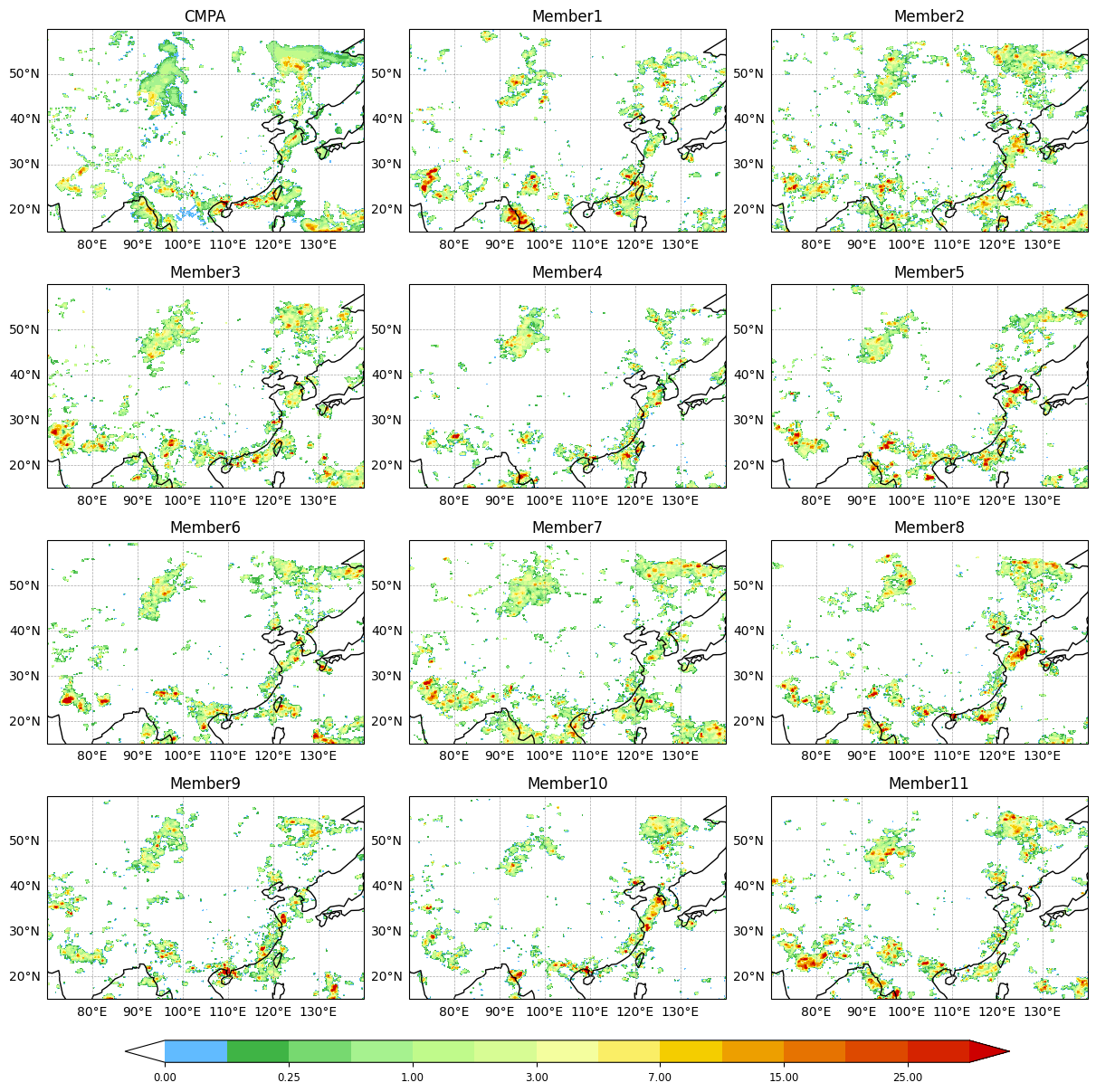}
    \caption{Precipitation of CMPA and all members generated by our ensemble system on 2021-08-01 01:00.}
    \label{fig:case1_all}
\end{figure}

The results begin with a case study that effectively demonstrates the model’s inference process. First, a deterministic precipitation forecast is generated at a 0.25° grid resolution using the deterministic model. As shown in Figure~\ref{fig:case1}, the overall precipitation distribution captures the spatial patterns of the target quite well, although the predicted precipitation intensity is lower than the target.

Next, Gaussian noise is sampled from the latent space and used as input for the denoising process, conditioned on both the atmospheric state and the deterministic precipitation forecast. The denoising process consists of 300 steps, ultimately producing the residual precipitation forecast at 0.05° grid resolution. By repeating this sampling process 11 times, we generate 11 distinct residual precipitation forecasts, effectively capturing the uncertainty at a finer scale. The spatial distribution of the residual precipitation also closely aligns with the target. Notably, this residual precipitation is derived from the difference between scaled CMPA precipitation and scaled ERA5 precipitation. The residual precipitation is then reverted from its normalized form, and when added to the deterministic precipitation forecast, it yields the final precipitation forecast. We generated the ensemble precipitation by applying probability matching \cite{ebert2001} across the 11 ensemble members. Using CMPA precipitation as a reference (which we consider to be more representative of true precipitation), it is evident that the ERA5 forecast overestimates the precipitation area while underestimating the precipitation intensity. The ensemble forecast, however, significantly improves upon these biases, providing better estimation for both the precipitation area and intensity.

To further illustrate this, we also present the individual precipitation forecasts from the 11 ensemble members (Figure~\ref{fig:case1_all}). While each member exhibits slight variations, they share a broadly consistent spatial distribution, demonstrating that our forecasting method is both influenced by large-scale atmospheric conditions and capable of representing small-scale uncertainties.

\subsubsection{Rank histogram and CDF}
\begin{figure}[ht]
    \centering
    \noindent\includegraphics[width=12cm]{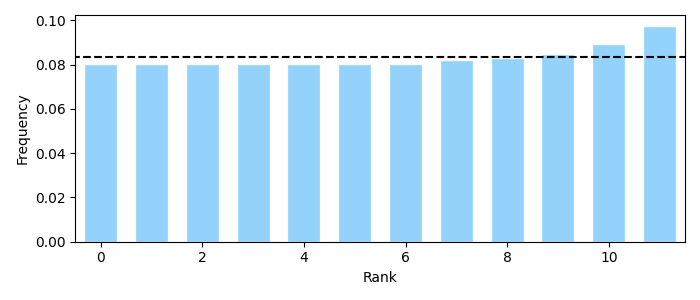}
    \caption{Rank Histogram. Blue bars showing the probability of the observed precipitation’s rank among the ensemble members. The black dashed line  The black dashed line represents the case where the true value is equally likely to fall between any two ensemble values.}
    \label{fig:rank}
\end{figure}

In addition to the case study, we further evaluate the performance of the ensemble forecast using two statistical tools: the rank histogram and the cumulative distribution function (CDF). These methods help us understand the reliability of the ensemble system and how well it captures the precipitation distribution compared to observed data.
The rank histogram provides insights into the reliability and spread of the ensemble forecasts by comparing the ensemble’s forecast ranks against observations. As shown in Figure~\ref{fig:rank}, the rank histogram exhibits a nearly uniform distribution, with only slightly elevated tails. This indicates that in some cases, the system may slightly overestimate precipitation. Despite this minor bias, the overall uniformity of the rank histogram suggests that the ensemble captures the inherent uncertainty in the atmospheric state well, making it valuable for operational forecasting.

\begin{figure}[ht]
    \centering
    \noindent\includegraphics[width=8cm]{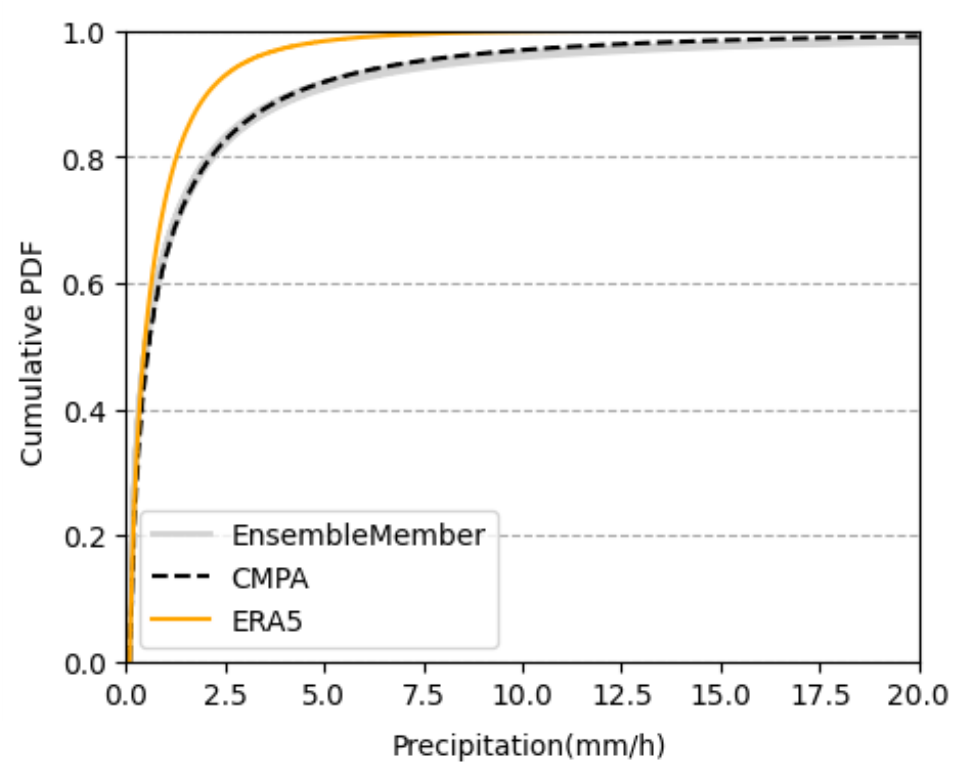}
    \caption{Cumulative PDF of  precipitation. The yellow line represents ERA5 precipitation, the black dashed line represents CMPA precipitation, and the gray curves represent the 11 ensemble members’ forecasts.}
    \label{fig:cdf}
\end{figure}

The CDF offers a complementary view to the rank histogram by comparing the ensemble forecast distribution to the reference data (CMPA) over a specific period and region. From August 9 to 12, 2021, southern China experienced continuous heavy precipitation, leading to severe flooding. We analyzed the precipitation distribution during this period within the region of 17\textdegree{}N–35\textdegree{}N and 100\textdegree{}E–125\textdegree{}E. As shown in Figure~\ref{fig:cdf}, the frequency distribution of precipitation reveals that all of the high-resolution ensemble member aligns more closely with the CMPA precipitation distribution than the ERA5 dataset, while ERA5 tends to overestimate light precipitation events and underestimate heavy ones. This is a key improvement, as correctly capturing the distribution of heavy precipitation events is crucial for flood risk management and disaster response. 
\subsection{Real-time ensemble forecasting}
We also develop a real-time high-resolution precipitation forecasting system using publicly available ECMWF data as input for our model. Since the ECMWF real-time open dataset provides forecasts at 3-hour intervals, we employ the Pangu-Weather model to perform 2-hour extrapolations, allowing us to obtain hourly atmospheric forecasts. Based on this data, we construct a high-resolution ensemble precipitation forecasting system for the next 120 hours.

To evaluate the system’s performance in real-world scenarios, we compare the forecast results against CMPA precipitation as a reference. We assess the Critical Success Index (CSI), Probability of Detection (POD), and False Alarm Ratio (FAR) metrics. The forecast initialization times are at 12:00 UTC for each day from August 1 to August 31, 2021, with a forecast range of 120 hours.

The CSI scores exhibit a periodic fluctuation with a 3-hour cycle, likely due to the inference predictions introduced by the Pangu model. Overall, for forecast lead times within 24 hours, for a threshold of 0.1 mm/h, the average CSI score across the 11 ensemble members ranges between 0.2 and 0.35, while for 2 mm/h, the CSI score is approximately 0.1. As expected, higher thresholds lead to lower CSI scores, reflecting the increasing difficulty in accurately predicting heavy precipitation events (this may be attributed to the relative rarity of high-intensity rainfall events, which makes them more challenging to predict).

As the forecast lead time increases, the FAR gradually rises, and the CSI shows a decreasing trend. However, even by lead time of 5 days, the CSI for the 0.1 mm/h threshold remains at a reasonable level of 0.15–0.3, while for 2 mm/h, it stays around 0.05. We also plot the CSI scores for the best-performing ensemble member at each forecast time. These results indicate that with the adoption of a more advanced post-processing or ensemble integration method, forecast accuracy could be further improved by optimally selecting or weighting the best ensemble members.
\begin{figure}[ht]
    \centering
    \noindent\includegraphics[width=12cm]{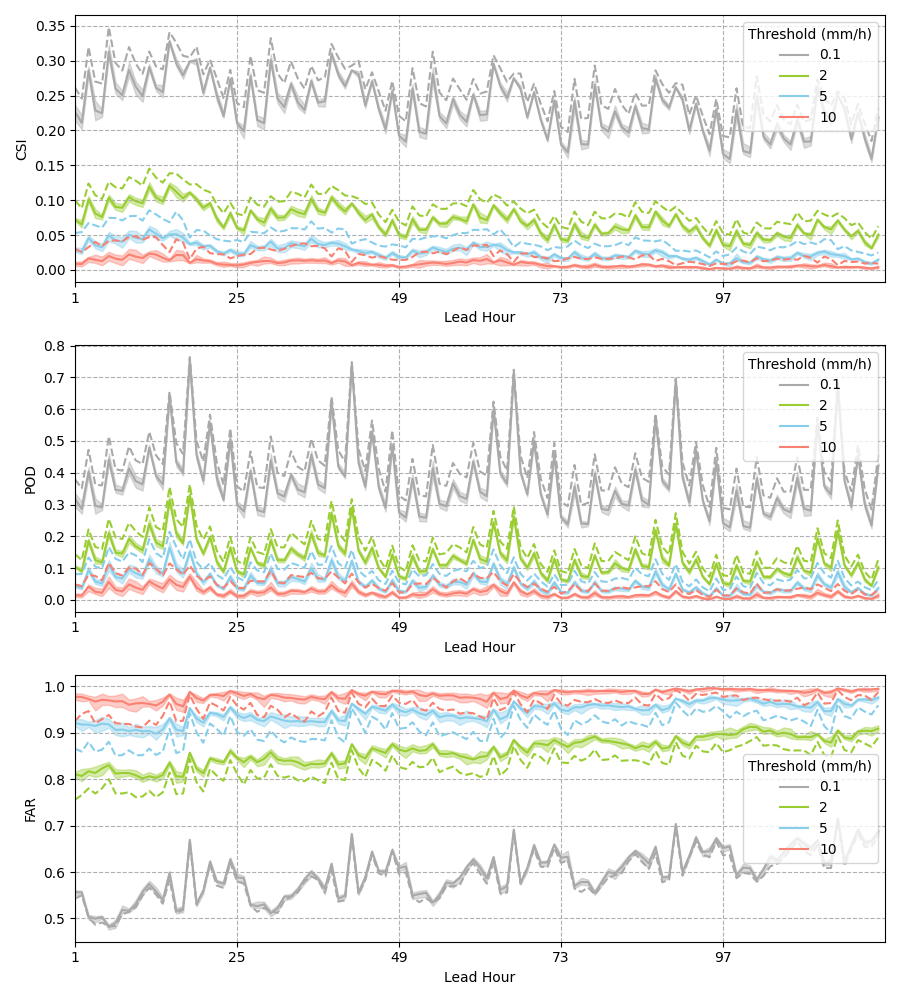}
    \caption{The average CSI, POD, and FAR from the real-time forecast system for August 2021, over 120-hour forecasts issued at 12 UTC each day. Shaded areas show the range of values across individual ensemble members; the solid line shows average value across all members; the dashed line indicates the metrics of the best-performing member.}
    \label{fig:online_score}
\end{figure}

\section{Conclusion}
This study proposes a physics-inspired deep learning framework that combines deterministic and probabilistic modeling approaches to significantly improve the accuracy and spatial resolution of high-resolution precipitation forecasting. The model is trained using ERA5 and CMPA datasets, with atmospheric variables such as temperature, geopotential, humidity, and wind as input, and produces high-resolution precipitation forecasts at 0.05° resolution. Built on a 3D Swin Transformer backbone, the deterministic model significantly improves diagnostic accuracy, especially for moderate to heavy rainfall, through improvements in patch embedding/recovery, loss function, and the incorporation of geographic and temporal features. With atmospheric state and meso-scale precipitation as a constraint, a conditional diffusion model in latent space is applied to capture uncertainties at convective scales after compressing the data using VAE. During inference, ensemble forecasts are generated by random sampling of Gaussian noise in the latent space. Our evaluation shows that the constructed ensemble forecast system is nearly unbiased, and its precipitation distribution is closer to CMPA compared to ERA5. Additionally, we developed a real-time forecast system based on ECMWF’s public dataset, which demonstrated robust performance in terms of CSI scores.

Overall, this study presents a novel approach to high-resolution precipitation forecasting. Due to data limitations, when transferring the model trained on ERA5 to ECMWF data, we didn't apply fine-tuning beyond adjusting the precipitation data scale in this study. Despite the high homogeneity between the two datasets, differences in resolution and other factors still introduce some discrepancies. In the future, as more data becomes available, we will apply model fine-tuning to further enhance the real-time forecast system’s performance.

\section*{Data Availability}
The ERA5 dataset used in this study was accessed from the Google Cloud public data archive, which can be found at https://console.cloud.google.com/storage/browser/gcp-public-data-arco-era5. The ECWWF open real-time forecast with 0.25\textdegree{} resolution is available at https://console.cloud.google.com/marketplace/product/bigquery-public-data/open-data-ecmwf; The CMPA precipitation dataset was provided by the China Meteorological Administration’s National Meteorological Information Center.

\section*{Code Availability}

\acknowledgments
We would like to thank the researchers at ECMWF for their invaluable work in offering the ERA5 dataset and open operational data; and the colleagues for discussions and suggestions at ColorfulClouds Tech.

%
%

\bibliography{agusample}

%
%
%
%
%
\clearpage
\appendix
\section{Evaluation metrics}
\subsection{POD, FAR and CSI}
POD (Probability of Detection), FAR (False Alarm Ratio), and CSI (Critical Success Index) are widely used in forecast evaluations. The POD measures the fraction of observed events that were correctly predicted; the FAR measures the fraction of predicted events that did not occur; and the CSI quantifies the success of forecasts for hit rates.

\begin{equation}
    POD = \frac{Hits}{Hits + Misses}
\end{equation}

\begin{equation}
    FAR = \frac{False Alarms}{Hits + False Alarms}
\end{equation}

\begin{equation}
    CSI = \frac{Hits}{Hits + Misses + False Alarms}
\end{equation}
\subsection{Rank histogram}
The rank histogram \cite{talagrand1999} shows the position of the ground truth within the ensemble distribution. For each time step and grid cell, rank the N ensemble members from lowest to highest for each observation and identify the rank of the observation with respect to the forecasts. Rank histograms can help detect under or over-dispersion \cite{hamill2001, wilks2011rank}. Under-dispersion causes the ground truth to fall near or outside the outer ensemble limits, creating a $\cup$-shaped histogram. Over-dispersion places the ground truth near the center, resulting in a $\cap$-shaped histogram. Peaks at either end may indicate bias in predictions. A flat rank histogram indicates unbiased ensemble forecasts, with the true value equally likely to fall between any two ensemble values.
\end{document}